\documentclass{article}

\usepackage{PRIMEarxiv}
\usepackage{placeins}

\usepackage[utf8]{inputenc} 
\usepackage[T1]{fontenc}    
\usepackage{hyperref}       
\usepackage{url}            
\usepackage{booktabs}       
\usepackage{amsfonts}       
\usepackage{nicefrac}       
\usepackage{microtype}      
\usepackage{lipsum}
\usepackage{fancyhdr}       
\usepackage{graphicx}       
\graphicspath{{media/}}     
\usepackage{subfig}
\usepackage{soul}
\usepackage{caption}
\usepackage{siunitx}
\usepackage{authblk}
\usepackage{csvsimple}

\title{Web Content Filtering through knowledge distillation of Large Language Models}

\author{Tamás Vörös}
\author{Sean Paul Bergeron}
\author{Konstantin Berlin}

\affil{Sophos Inc. \authorcr
  \{\tt tamas.voros, sean.bergeron, konstantin.berlin\}@sophos.com}

\begin{document}
\maketitle

\begin{abstract}

We introduce a state-of-the-art approach for URL categorization that leverages the power of Large Language Models (LLMs) to address the primary objectives of web content filtering: safeguarding organizations from legal and ethical risks, limiting access to high-risk or suspicious websites, and fostering a secure and professional work environment. Our method utilizes LLMs to generate accurate classifications and then employs established knowledge distillation techniques to create smaller, more specialized student models tailored for web content filtering. Distillation results in a student model with a 9\% accuracy rate improvement in classifying websites, sourced from customer telemetry data collected by a large security vendor, into 30 distinct content categories based on their URLs, surpassing the current state-of-the-art approach. Our student model matches the performance of the teacher LLM with 175 times less parameters, allowing the model to be used for in-line scanning of large volumes of URLs, and requires 3 orders of magnitude less manually labeled training data than the current state-of-the-art approach. Depending on the specific use case, the output generated by our approach can either be directly returned or employed as a pre-filter for more resource-intensive operations involving website images or HTML.
\end{abstract}


\section{Introduction}
Web content filtering is crucial for maintaining network security and regulatory compliance in organizations\cite{webcontentfiltering, baishya2019webcontent}. The aim of a web content filtering system is to prevent employees from accessing inappropriate content that violates regulatory requirements or company policies, and by filtering out high-risk content categories, such as pornography and weapons, it helps to avoid legal liability, reduces the risk of legal or ethical issues arising from exposure to unsuitable content, and promotes a professional work environment. Unlike security classification, which detects hosted malware and phishing attacks, content filtering models address a more general problem that is independent of the attack mechanism. In this work, we address the problem of web content categorization.

\begin{figure}[!htbp]
    \centering
    \includegraphics[width=0.55\textwidth]{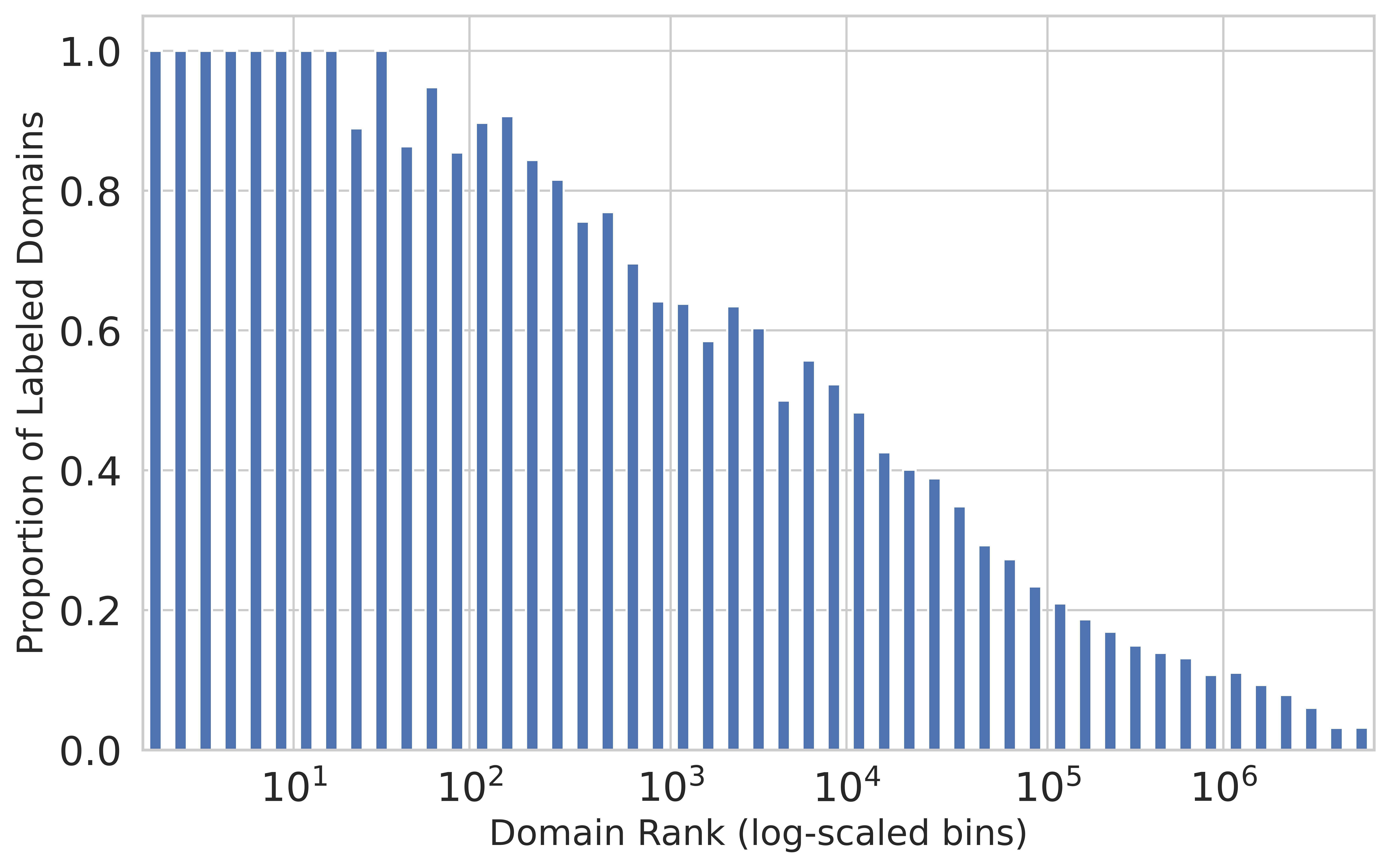}
    \caption{Labeling drop-off. The plot visualizes the proportion of analyst-labeled domains across different popularity levels of a large security vendor. The domain popularity is represented on the $x$-axis using a logarithmic scale, where higher values indicate more popular domains. Each bar in the plot corresponds to a specific popularity bin, with the height of the bar illustrating the proportion of labeled domains within that bin.}
    \label{fig:coverage}
\end{figure}

Traditional approaches to website categorization have relied upon creating and maintaining domain-to-category mappings, which are lists of domains grouped by their manually assigned categories \cite{sheng2009blacklist}. A natural extension to list-based URL categorization is to enhance them with signatures created by analysts, that would generalize better than exact string matching. In the case of web content filtering, the most straightforward signature-based approach is to propagate labels based on domains and subdomains, although more complex rules may be applied \cite{chen2020improving, huang2010mitigate, haruta2019novel}. An example of this kind of label propagation is to maintain a list of known domains with predetermined labels, such as labeling ``online-shop.com'' as an e-commerce site and ``news-site.com'' as a news site. All URLs under these domains inherit the label. For instance, any URL under ``online-shop.com'' such as ``online-shop.com/products/clothing'', ``online-shop.com/products/electronics'', and ``online-shop.com/cart'' can be labeled as e-commerce. Similarly, any URL under ``news-site.com'', such as ``news-site.com/politics'', ``news-site.com/technology'', and ``news-site.com/entertainment'' can be labeled as news. In the manuscript, we focus on domain label propagation signatures for acquiring ground truth for the sake of simplicity, but it could be trivially extended to longest prefix matching of the URL for ambiguous websites.  To provide comprehensive customer telemetry coverage for organizations, one of the most resource-effective manual methods involves ranking domains by frequency and labeling them in descending order. This approach maximizes coverage by prioritizing the labeling of a single domain. 
As new websites emerge daily and with over a billion existing websites, maintaining and scaling signature approaches manually for the long tail has become increasingly challenging. This necessitates the integration of machine learning into the classification pipeline\cite{ma2009beyond, saxe2017expose, maneriker2021urltran}. 
Figure \ref{fig:coverage} illustrates the telemetry coverage of a large security vendor, with the space above the bars representing the infrequently seen long tail distribution of domains not already covered by domain labeling and label propagation signatures. 

Maintaining domain-to-category mapping lists and extending them with signatures remains critical in the early stages of security pipelines \cite{apruzzese2022pipeline}. These labels serve as initial shortcuts in the filtering pipeline to prevent catastrophic false positives, and provide low latency on more commonly seen websites. Websites like 'stackoverflow.com' are well-known and need not be evaluated by a model whereas a potential false positive would translate to a negative impact on the productivity of an organization. 
In this work, we focus our evaluations on the long tail of the distribution, which aligns with actual deployment scenarios and emphasizes the need for machine learning to address the challenges associated with classifying this ever-growing subset of domains.

In addition to acting as a pre-filter, domain-to-category mapping lists and label propagation signatures are often used to create the training sets for machine learning models. However, machine learning algorithms tend to memorize patterns rather than understand underlying concepts \cite{zhang2022generalize, bhargava2021generalization}, thus learning from already labeled URLs is insufficient for accurate content classification in the long tail of the URL distribution. A model whose parameters are configured to memorize the head of the distribution is undesired as signatures already cover such domains without risking false positives. Therefore, our objective is to identify models with superior generalization capabilities for out-of-distribution samples.

For unknown or new domains, the model must infer a description from the URL. It is useful to view URL classification, especially for web content filtering, as a natural language processing task, considering URLs as semi-sentences. For a fair amount of our categories, the URL will frequently have explicit words to advertise its content, specifically semantically related keywords for the given category. For example, a site selling weapons will often contain keywords such as ``armaments'' or ``glock'', or ``gun''. The current state-of-the-art in URL detection and our chosen baseline,  URLTran \cite{maneriker2021urltran}, frames URL detection as a natural language processing task and fine-tunes a pre-trained BERT model \cite{devlin2018bert} to detect phishing URLs. The BERT model is an early example of the transformer architecture \cite{vaswani2017attention} which has since been refined and scaled, giving rise to large language models. Large language models (LLMs) are state-of-the-art on natural language tasks \cite{brown2020language}. LLMs are first pre-trained on large amounts of unlabeled textual data in a task-agnostic manner, learning a general understanding of language such as syntax and semantics \cite{brown2020language}. Once pre-trained LLMs can effectively generalize to new tasks upon fine-tuning or few-shot prompting with much smaller amounts of data \cite{brown2020language}. The amount of data needed for LLMs to generalize to new tasks is often several orders of magnitude less than the amount of data needed to fully train a smaller model.

Direct use of LLMs for URL content classification in production is prohibitive due to cost considerations at scale \cite{howmanywebsites}. Fine-tuning smaller LLMs that have lower inference costs results in a loss of performance. 
Through knowledge distillation \cite{hinton2015distilling}, the LLM-labeled long tail data enables a smaller student model to improve its performance while maintaining the necessary computational efficiency for production. Turc et al. \cite{teacherstudent} proposed an approach that utilizes knowledge distillation from the teacher's predictive distribution (soft labels) followed by supervised fine-tuning of the student model. In the domain of web content classification, we combine the steps of distillation and fine-tuning and our computationally efficient student matches the performance of the teacher model. Instead of a predictive distribution, we distill the teacher using hard labels. The student model has a low inference cost and is well-suited for the purposes of web content filtering in production.
 
The main contributions of this paper are as follows:

\begin{itemize}
\item We demonstrate that when fine-tuned on data labeled with domain propagation signatures, large language models outperform standard deep learning models by 9\% in terms of accuracy on the long tail categorization problem. \

\item We demonstrate that we can fine-tune a large language model using 10000 samples to achieve better performance than the current state-of-the-art approach trained on 10 million samples. \

\item We showcase the effective application of knowledge distillation from a fine-tuned LLM to boost the performance of a smaller, more computationally efficient model, specifically for web content filtering tasks. We attain performance levels comparable to the original LLM using a model that is 175 times smaller, decreasing from 770 million parameters to just 4 million. This reduction in size makes the model more suitable for production and enables practical deployment across various contexts, such as serving as a general pre-filter for all incoming network traffic in firewalls.

\item We propose a novel validation approach for the community to adopt, which more accurately assesses model performance in realistic scenarios where it works alongside a domain-to-category mapping list of ground truth labels, extended via domain label propagation signatures. In this setting, the model analysis focuses on labeling the long tail, focusing on a more relevant metric.

\end{itemize}

Our paper is structured as follows: In Section 1 we introduce the research problem and elucidate the motivation behind our proposed approach. In Section 2 we review relevant literature and prior work in the field. In Section 3, we provide a comprehensive description of our methodology, encompassing the dataset and experimental setups. In Section 4 we present our results, which include a comparison of our approach's performance against the current state-of-the-art, an analysis of the benefits of LLMs in terms of accuracy and sample efficiency, as well as an exploration of deployment challenges and our proposed solution utilizing knowledge distillation for more compact and computationally efficient models. Lastly, In Section 5 we conclude the paper, outlining potential avenues for future research in this domain.

\section{Related work}

Previous work in this field has primarily focused on security classification rather than content classification and filtering. Since machine learning approaches to security classification can be readily reformulated from binary classification to multi-class classification through modification of the last layer in the neural network, approaches to security classification are relevant to the task of content classification. We will compare and build upon security publications as they are better studied.

Early work on URL-only classification for phishing detection using manually derived feature sets employed both generic features and features meant to detect certain obfuscation techniques such as obfuscation of the host with another domain \cite{garera2007framework}. The features were divided into four groups: Page Based,
Domain Based, Type Based, and Word Based. The authors focused on manual feature engineering and only applied logistic regression as their classifier. A range of machine learning models, including Random Forests, Logistic Regression, Support Vector Machines, Naive Bayes, and Gradient Boosting, have been applied to detect phishing URLs using manually extracted feature sets \cite{ma2009beyond, oshingbesan2022detection}.
Feature sets may be entirely lexically derived such as the length of the URL, the number of digits in the primary domain, and the number of special characters in the path \cite{joshi2019using}. In addition to lexical features, domain-specific features such as the number of passive DNS changes or the remaining time of the SSL certificate may be incorporated \cite{hajaj2022less}. Manual features may also be extracted from the retrieved information of lookups (Whois, GSB Reporting, Google Ranking, and Selenium Rendering) \cite{abuadbba2022towards}.

The manual feature extraction approach is difficult to maintain as adversaries tend to adapt obfuscation methods to avoid detection so models have shifted to a featureless approach based on the raw string as input. Deep learning methods learn and then automatically extract the feature set from the raw URL during training. The use of automatically extracted features does not preclude the inclusion of manual features however as the optimal input combination of manual and automatic features can be optimized with genetic algorithms \cite{bu2022optimized, park2021evolutionary}.

Automatic feature extraction can be done on various levels of granularity starting at the character level. Saxe et al. \cite{saxe2017expose} encode a URL by replacing each character with its corresponding ID whereby features are extracted from the encoded URL with sequential embedding and convolutional layers. This approach outperformed a baseline which uses a manual feature set. Learning meaningful context-independent representations is difficult when using character-level tokenization as a character token doesn't carry the same meaning that a word does. More recent approaches like subword-level and word-level tokenization have been developed in natural language processing in order to make it easier for models to maintain semantic meaning in common subwords and learn more meaningful context-independent representations. 

The application of word-level tokenization to URL classification was first proposed by 
Le et al. \cite{le2018urlnet} who extracted both character-level and word-level features. Each feature set is fed through its own series of sequential embedding and convolutional layers before being fused. Tajaddodianfar et al. \cite{tajaddodianfar2020texception} expand on this approach by first training the word embeddings in an unsupervised manner via FastText \cite{joulin2016fasttext}. The word and character convolutional stems include several convolutional layers in parallel with dilated convolutions allowing the model to adaptively grow in depth and width, extracting N-grams of various lengths. In addition to using both character and word-level feature models, Bu et al.\cite{bu2021learning} apply a triplet network structure in order to address class imbalances and better learn the similarity between URLs. 

In addition, to feature set selection, the choice of model architecture plays a large role in the performance of a URL classification model. Transformers have achieved state-of-the-art results in many natural language processing tasks making them a good candidate for URL classification after fine-tuning or even custom pre-training \cite{rudd2020training, rudd2022transformers, maneriker2021urltran, chang2021research, shirazia2022towards}. In addition to the URL, a transformer can leverage tokenized features of the HTML \cite{hu2021phishing}. A URL classification system might employ different architectures in parallel, fusing the output of models with a convolutional architecture and a transformer architecture  \cite{wang2022tcurl}. Instead of fusing model outputs, a system may employ an ensemble of different architectures including Decision Trees, LSTMs, and transformers for URL classification \cite{venugopal2021detection}. 

Other architectures applied to URL classification include graphical networks \cite{ariyadasa2022combining, bilotphishgnn} and GANs \cite{kamran2021semi, geng2022effective}. AutoEncoders have proven useful against zero-day attacks \cite{bu2021deep}. In addition to the URL and HTML sequences, but beyond the scope of this paper, images of the webpage may be incorporated \cite{yuan2021malicious, liu2022inferring}. The task of classification may be reformulated by approaching detection from a reinforcement learning perspective \cite{lavie2022transferable} or from the perspective of thwarting an adversarial opponent \cite{peng2022crafting, kim2022phishing}.

The current state-of-the-art for URL-only classification for phishing detection, URLTran \cite{maneriker2021urltran}, utilizes the transformer architecture underpinning LLMs. Maneriker et al. fine-tune a pre-trained BERT model on Microsoft’s Edge and Internet Explorer production browsing
telemetry. Parallel to URLTRan is the Unified Text-to-Text Cybersecurity (UTS) model. Pal et al. \cite{pal2023exploring} train a multi-task encoder-decoder LLM on cybersecurity data that includes URL phishing detection. Although Pal et al. introduce LLMs to URL phishing detection, they do not explore the few shot capabilities of LLMs in the URL domain nor test the capabilities of LLMs at scale. Compared to URLTran, UTS does not consider a methodology which would allow its large model to be used in production and reports a lower F1 score on a random split compared to URLTran's evaluation on the industry standard time split. Therefore, URLTran will act as our baseline to which all of our results will be compared.

\section{Methodology}
In this section, we describe our methodology for collecting data and constructing training, validation, and test sets. We also explain our experimental setup and provide a detailed account of how we trained our model.

\subsection{Data}

We obtained our dataset from a large security vendor's customer telemetry data sourced from its firewall and endpoint products over a period spanning July 1, 2022 to December 23, 2022. 

We track 30 categories in our dataset. These categories were defined by a team of expert analysts to be representative of the most common internet content categories as well as the most impactful, which we define as the potential to impact productivity, the degree of liability for the organization, and the degree of associated ethical concerns. The categories include: ``Chat'', ``Games'', ``Shopping'', ``Sports'', ``News'', ``Job Search'', ``Search Engines'', ``Alcohol'', ``Gambling'', ``Weapons'', ``Porn'', ``Banking'', ``Business'', ``Education'', ``Entertainment'', ``Food and Dining'', ``Government'', ``Health and Medicine'', ``Motor Vehicles'', ``Peer to Peer'', ``Real Estate'', ``Religion'', ``Travel'', ``Translators'', ``Computer and Internet'', ``Hunting and Fishing'', ``Marijuana'', ``Radio and Audio Hosting'', ``Social Networking'', and ``Video Hosting''. The majority of websites in our dataset belong to categories such as  ``Computer and Internet'',  ``Search Engines'', and  ``Business'', while niche categories such as ``Hunting and Fishing'' and ``Marijuana'' have fewer instances. Figure \ref{fig:category_distribution} shows the distribution of categories in our dataset. We define our categorization task as a closed-world problem, meaning every URL belongs to one of the 30 categories. It's important to note that, due to limitations in the domain-to-category database, we only consider a single category per URL, even though some pages may realistically have multiple category labels.

\subsection{Training sets}

To construct our training dataset which spans the period from July 1, 2022 to August 19, 2022, we uniformly sampled 10 million distinct URLs, out of the billions of URL lookups, that have been labeled using a domain-to-category mapping database with label propagation. Additionally, we sampled 10 million URLs from this period that did not correspond to a signature (unlabeled). The unlabeled URLs were set aside for training augmentation purposes. 

\subsection{Validation and Test sets}

We sampled an evaluation dataset spanning from August 19, 2022 to December 23, 2022 and divided it into two validation and test sets to assess our model's performance in different scenarios. The validation sets were based on data first seen between August 19, 2022 and November 24, 2022 while the test sets included data first seen between November 24, 2022 and December 23, 2022.

We created a domain and time split to simulate a long tail deployment setting. We separated the data based on the first-seen time of the URL and the first-seen time of the URL's domain. The first-seen time of a domain refers to the earliest instance of a URL from that domain. Meaning, there is no domain overlap between the training, validation, and test sets. This approach allowed us to better approximate the unlabeled part of the telemetry.

To compare our results with the industry-standard evaluation methodology we also created a time split. This split was sampled from the same time span as the domain and time split but without the constraint of dividing based on the domain's first-seen time.

For the domain and time split, the validation set was comprised of 79,313 unique URLs from 30,897 unique domains, with a maximum of 5 URLs per domain. The test set included 110,624 unique URLs from 43,996 domains. For the time split, we sampled 183,935 URLs from 62,961 domains.

To compare the various splits, we display the most common domains and their frequencies for the labeled training data, both test splits, and the unlabeled training data in Table \ref{tab:domains_dist}. The labeled training set and the time split are dominated by common domains such as ``google.com''. The domain and time split is most similar to the unlabeled long tail of the data where the desired value of machine learning resides. Infrequent domains.

\begin{table}[h!]
    \centering
    \resizebox{\textwidth}{!}{\begin{tabular}{l|r|l|r|l|r|l|r}
        \toprule
        \multicolumn{2}{c|}{Training Set} & \multicolumn{2}{c|}{Time Split Test Set} & \multicolumn{2}{c|}{Domain and Time Split Test Set} & \multicolumn{2}{c}{Unlabeled Data} \\
        \midrule
        \textbf{Domain} & \textbf{Frequency \%} & \textbf{Domain} & \textbf{Frequency \%} & \textbf{Domain} & \textbf{Frequency \%} & \textbf{Domain} & \textbf{Frequency \%} \\
        \midrule
        \begin{filecontents*}{frequencies.csv}
,domainRandom,randomval,domainTime,timeval,domainDomain,domainval,domainUnlabeled,unlabeled
0,google.com,20,google.com,33,tomcleaneraddon.com,<1,wymondhamcollege.org,2
1,microsoft.com,6,microsoft.com,4,ammdx.com,<1,mfa.cloud,1
2,googleapis.com,5,gstatic.com,2,ogp.me,<1,dimmittisd.net,<1
3,cedexis-radar.net,3,googlesyndication.com,2,officested.com,<1,qq.com.cn,<1
4,gvt1.com,3,doubleclick.net,2,dimensionu.com,<1,gnsmat.co.uk,<1
5,facebook.com,2,msn.com,2,shreemaruti.com,<1,headlandentertainment.com,<1
6,zeotap.com,2,googleusercontent.com,2,wbe-eindhoven.nl,<1,murray.edu,<1
7,youtube.com,1,youtube.com,1,vapornodes.finance,<1,pcbid.top,<1
8,amazonaws.com,1,amazonaws.com,1,trendingtrck.com,<1,stoughtonwi.com,<1
9,sharepoint.com,1,cloudfront.net,1,bluedrop360.com,<1,aveha.com,<1
\end{filecontents*}
\csvreader[head to column names]{frequencies.csv}{}
        {\domainRandom & \randomval & \domainTime & \timeval & \domainDomain & \domainval & \domainUnlabeled & \unlabeled \\}
        
    \end{tabular}}
    \caption{Domains and their frequencies in the train and test sets.}
    \label{tab:domains_dist}
\end{table}

\begin{figure}[!htbp]
    \centering
    \includegraphics[width=0.55\textwidth]{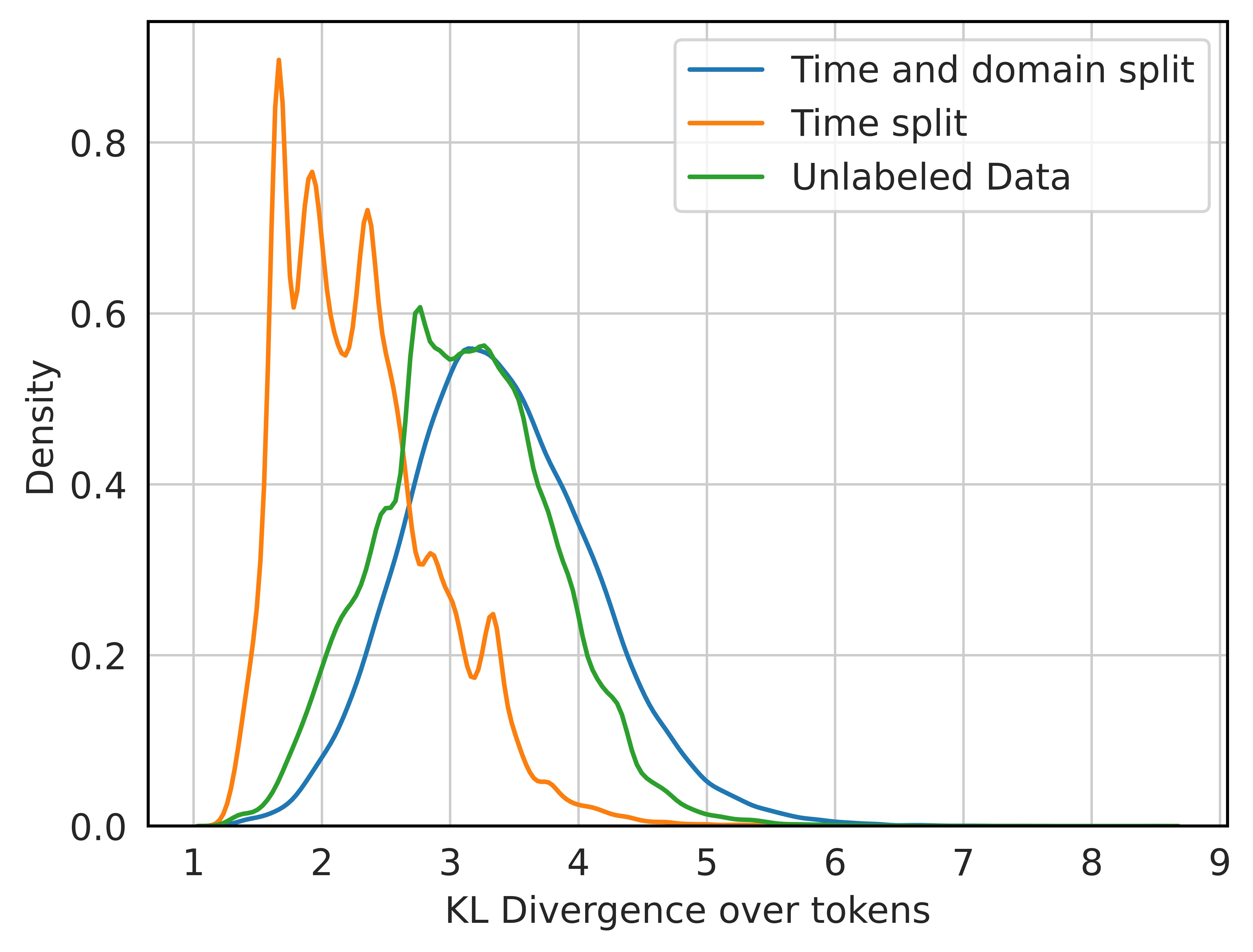}
    \caption{KL Divergence over BERT tokens. The $x$-axis represents the possible range of KL divergence values over BERT tokens, while the $y$-axis represents the estimated probability density of these values. The plot quantifies the difference between validation split URLs as compared to the training set token distribution, with higher KL divergence values indicating greater differences between the BERT token distributions of the base training set and validation split.}
    \label{fig:token_kl}
\end{figure}

To quantitatively assess the disparities between the domain distributions of the time split and the domain and time split, which more closely models the long tail, we employed the Kullback-Leibler (KL) divergence as a metric for measuring the dissimilarity between token distributions. The KL divergence values were calculated between each validation split and the training dataset as the reference. We tokenized all the URLs in the training dataset using BERT tokenization and then combined all of the tokens to define the distribution of the base training dataset. We tokenized all the URLs in both validation splits using BERT tokenization and each token sequence was converted into probability distributions by computing normalized histograms. The KL divergence between the token probability distribution of each URL and the reference distribution was then determined using the entropy function. Figure \ref{fig:token_kl} illustrates that the token distribution of the domain and time split displays substantially higher KL divergence values from the reference compared to the time split. This observation highlights the distinct nature of the domain distributions in the two validation splits and the similarity between the unlabeled part of the customer telemetry and the domain and time split.

\subsection{Experiments}

The primary objective of our experiments was to identify the best-performing model in terms of accuracy on our dataset, while using as few training labels and being as small as possible. To achieve this, we varied the training set size as a hyperparameter for each LLM, compact model, and the baseline. We explored training set sizes ranging from few-shot to large-scale learning, increasing the sample size from 10 samples per category to 5 million total samples, growing by an order of magnitude at each step. For a given sample step size $N$, the exact samples per category were determined by the minimum of $N$ and the total labeled instances in that category.

Our next goal was to refine the top-performing large language model (LLM) configuration into a more compact student model. We achieved this by using labels generated by the best-performing LLM to train smaller models.

We labeled 10 million unlabeled URLs from our dataset using the best-performing LLM, utilizing them as hard labels. This resulted in a total of 20 million training set with the 10 million signature-labeled base training set and an additional 10 million labels generated by the LLM. We then investigated the impact of combining these labels using various mixing ratios of labeled samples from the base training set and LLM-labeled samples. Each compact student model and baseline were trained on a variety of dataset configurations, each containing a total of 10 million samples.

We began with a 10-million base training set, incorporating LLM-generated labels at 0.0, 0.25, 0.50, 0.75, and 1.0 ratios. The 0.0 ratio used only the base training set, while the 0.25 ratio included 7.5 million base URLs and 2.5 million LLM-generated. At 0.5, the sources were evenly split with 5 million each. The 0.75 ratio contained 2.5 million base and 7.5 million LLM URLs, and the 1.0 ratio relied solely on LLM-generated labels. By varying the mixing ratios, we were able to assess the effectiveness of our knowledge distillation process and compare the contributions of LLM-generated labels to simply using signature-generated labels.

We trained and compared the performance of five models: BERT-based URLTran as the baseline\cite{maneriker2021urltran}, which demonstrated state-of-the-art performance for URL classification, eXpose \cite{saxe2017expose} and BERTiny \cite{bhargava2021bertiny} as the student models, and T5 Large \cite{raffel2020exploring} and GPT-3 Babbage \cite{brown2020language} as the teacher models. The size configurations of our teacher models were limited by budgetary constraints, precluding larger configurations such as GPT-3 Davinci and T5-11B. Our student models were chosen for the following reasons: BERTiny is the smallest pre-trained configuration of the baseline and the inclusion of eXpose allows us to demonstrate the improvements of the transformer architecture over convolutional models for natural language tasks, specifically web content categorization. Unless otherwise noted, all experiments were evaluated on the test set of both validation splits. The GPT-3 Babbage model was not fine-tuned on 5 million samples due to cost considerations. 

\subsection{Training}

For all models, we pre-processed the data by splitting at the first occurrence of the ``?'' character and removing the query parameters. The query is assumed to be noisy and without any meaningful information. All URLs were truncated to a fixed length of 128 characters as we have seen no improvement in further increasing the size. The base pre-trained models and tokenizers for all T5 Large, BERT, and BERTiny configurations were the HuggingFace defaults \cite{wolf2020transformers}. 

For all reported T5 configurations, we fine-tuned all weights of a pre-trained T5 Large model using the Adafactor optimizer \cite{shazeer2018adafactor}. Early stopping was applied by monitoring performance on the validation set of the domain and time split. For all reported GPT-3 configurations, we fine-tuned the Babbage model using the OpenAI API. 

T5 and GPT-3 are generative models that can utilize semantic relationships between class labels and keywords in a URL for making predictions. Consequently, we employed literal class labels as our prediction target. When reporting aggregate metrics, out-of-vocabulary (OOV) predictions are not considered as a separate class, and they were considered as misclassfication for every class. Additionally, any unlabeled data for which LLM generates an OOV prediction is excluded from the distillation process. 

For GPT3 the temperature was set to 0 to ensure deterministic results upon inference. The logit bias for tokens associated with the class labels were set to 100 to ensure exclusive selection of expected tokens. Finally, the stop token was set to the stop sequence seen during training. 

For the student models, we trained a 1D convolutional eXpose model and fine-tuned all weights of a pre-trained BERTiny model. We fine-tuned all weights of a pre-trained BERT model to reproduce the architecture of URLTran as our baseline. No custom vocabulary was created for the BERT-based models. Hyperparameter configurations for T5, BERTiny, BERT, and eXpose may be found in Tables \ref{tab:t5_hyperparameters}, \ref{tab:bertiny_hyperparameters}, \ref{tab:bert_hyperparameters}, and \ref{tab:eXpose_hyperparameters} respectively.

\section{Results}

In this section, we present the key findings and results of our two sets of experiments. We report the results in terms of accuracy, with additional metrics for both experiments provided in the Appendix.

\begin{figure}[htbp]
\centering
\subfloat[Scaling results]{\label{fig:results_scaling}\includegraphics[width=0.495\textwidth]{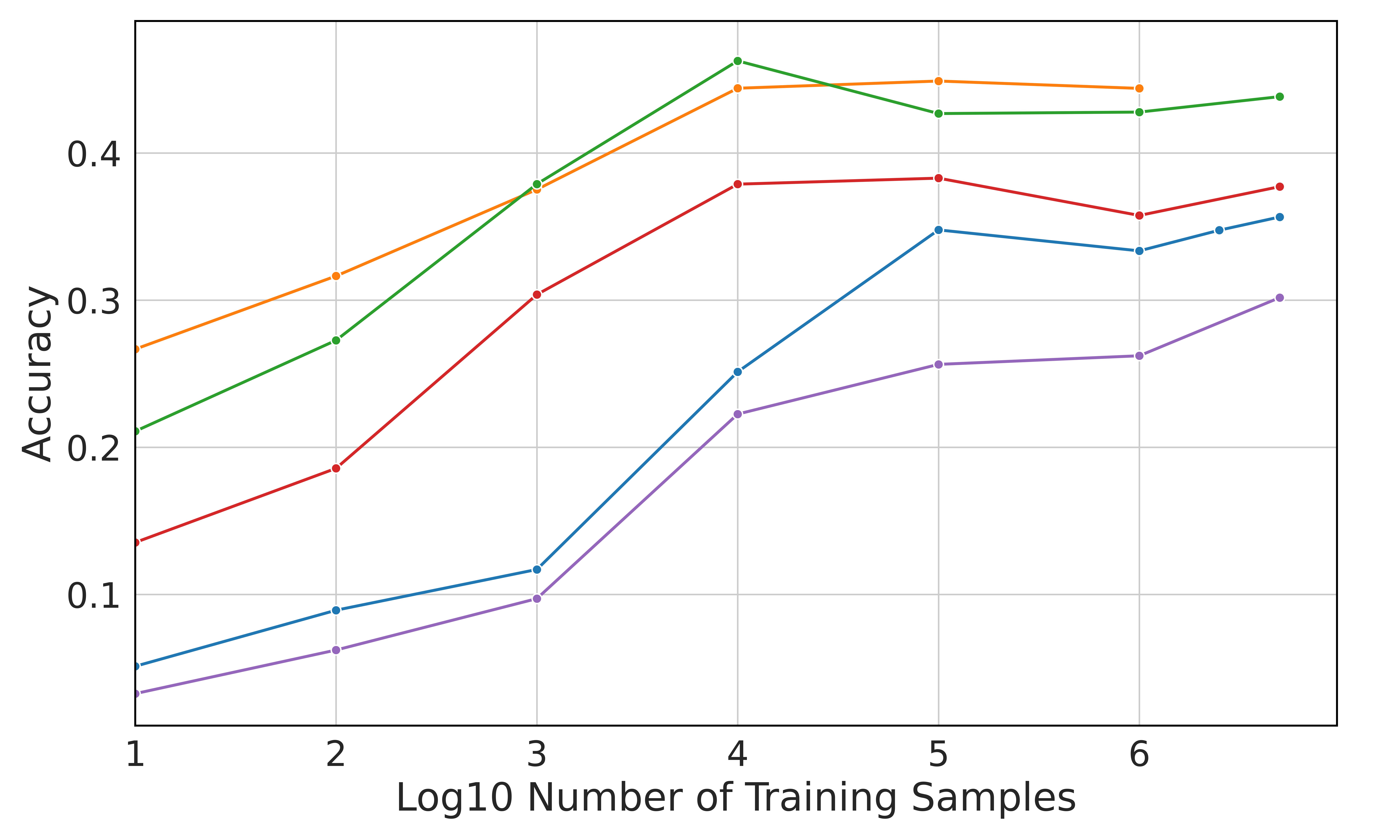}}
\hspace{0.005\textwidth} 
\subfloat[Augmentation results]{\label{fig:results_augmentation}\includegraphics[width=0.495\textwidth]{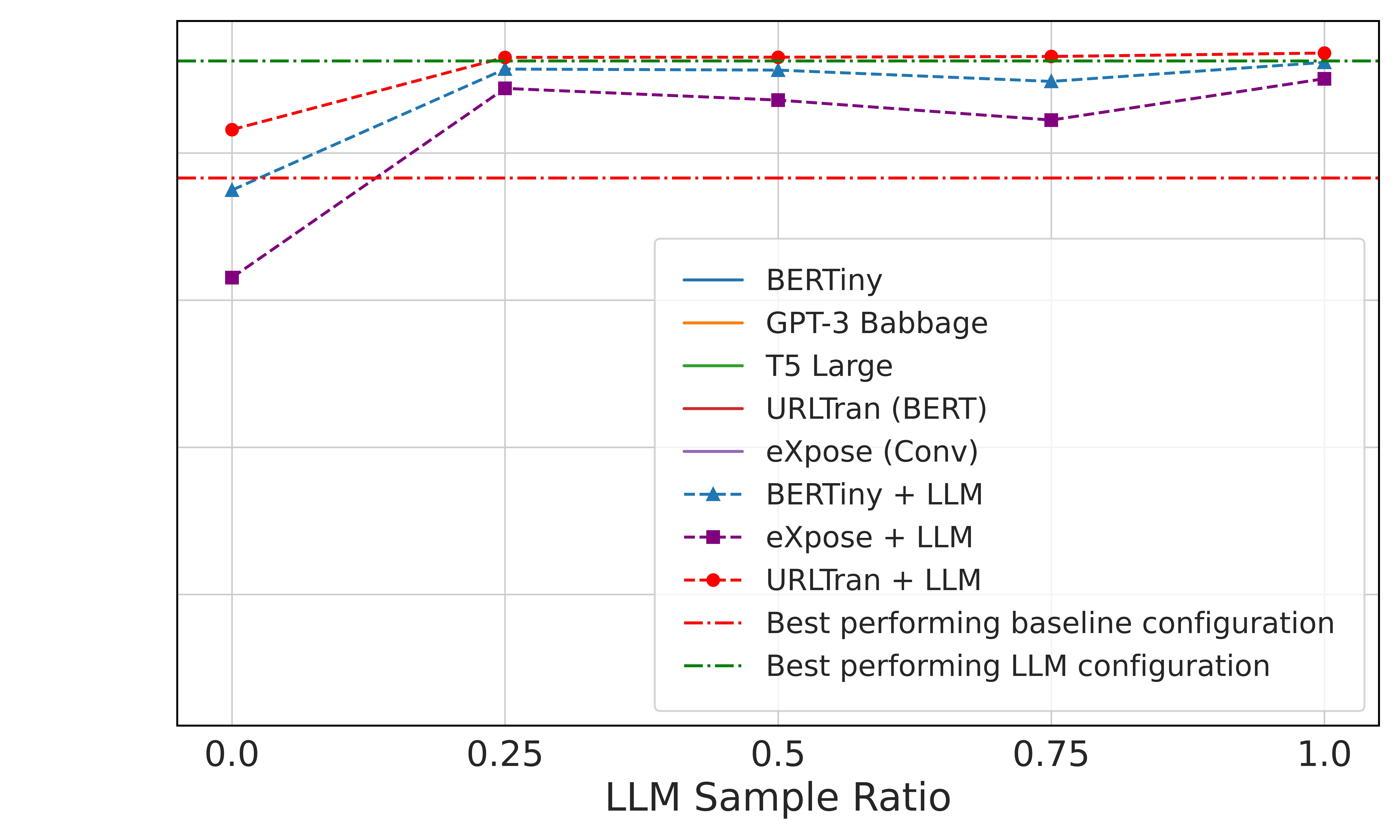}}
\caption{The results for scaling and augmentation are dependent on the domain and time split. Figure (a) illustrates the performance of various models in relation to the logarithm of training sample size. In Figure (b), we compare the top-performing LLM and baseline configurations with the performance of different student models as a function of the mixing ratio for LLM-generated labels. The GPT-3 Babbage model was not fine-tuned on 5 million samples due to cost considerations} %
\label{fig:results}
\end{figure}

The performance of various models as a function of the log of the training sample counts is displayed in Figure \ref{fig:results_scaling}. The top-scoring configuration for each model is detailed in Table \ref{tab:main-body-model-performance-domain}.
On the domain and time split, the best performing model, T5 Large, achieves 46.3\% accuracy after being fine-tuned on 10,000 samples. GPT-3 Babbage attains 44.4\% accuracy after fine-tuning on 10,000 samples. Both LLMs surpass the best baseline configuration, which achieves 38.3\% accuracy. BERTiny and eXpose achieve 35.7\% and 30.2\% accuracy, respectively, when trained on 5 million samples. 

On the time split, eXpose achieves 92.8\% accuracy when trained on 5 million samples. BERTiny, fine-tuned on 5 million samples, attains 97.6\% accuracy. The best configurations for the baseline, GPT-3 Babbage, and T5 Large achieve 97.1\%, 98.14\%, and 97.5\% accuracy, respectively. Additional metrics for the time split are reported in Table \ref{tab:model-performance-time}, and for the domain and time split in Table \ref{tab:model-performance-domain} for all experiments.

On the domain and time split, the best performance was achieved with T5 on 10,000 training samples, so we selected it as our teacher model. For the domain and time split, we found the best ratio to be 1.0, where every training sample had 10 million previously unlabeled URLs labeled by T5. Training eXpose on all of them increases the accuracy from 31.5\% to 45\%. Fine-tuning BERTiny on all 10 million LLM labels, compared to the 10 million base training set, improves the accuracy from 37.5\% to 46.2\%. Finally, fine-tuning URLTran on all 10 million LLM labels, compared to the 10 million base training set, raises the accuracy from 41.6\% to 46.8\%.

For the traditional time split, the augmentation at a ratio of 0.75 also increased the performance, albeit marginally.

The performance of the students and the baseline trained via knowledge distillation is shown in the augmentation plot of Figure \ref{fig:results_augmentation} as a function of the LLM label ratio in the training data. The top-scoring configuration for each model is detailed in Table \ref{tab:main-body-model-performance-domain}.

\begin{table}[htbp]
\small
\centering
\caption{The best performing model configurations are presented, including the distilled versions of the students and baseline. The accuracy for each model's top configuration is displayed, along with the model's parameter count relative to the best performing LLM. For the Time and Domain split, the LLM label ratios correspond to 1.0, and for the Time split, the ratio is 0.75, as these were consistently the best across the models. More detailed results can be found in Table \ref{tab:model-performance-domain} and Table \ref{tab:model-performance-time}.}
\label{tab:main-body-model-performance-domain}
\resizebox{\textwidth}{!}{%
\begin{filecontents*}{main_table.csv}
model,n_samples_int,accuracy_domain,macro_f1,macro_recall,macro_precision,weighted_f1,weighted_recall,weighted_precision,percent,accuracy_time, parameter_count
eXpose (Conv),$5 \times 10^6$,0.30,0.15,0.14,0.23,0.29,0.30,0.34,,0.93, 3.3
BERTiny,$5 \times 10^6$,0.36,0.24,0.22,0.30,0.36,0.36,0.38,,0.97, 4.4
URLTran (BERT),$1 \times 10^5$,0.38,0.25,0.22,0.33,0.38,0.38,0.41,,0.97, 110
T5 Large,$1 \times 10^4$,\textbf{0.46},0.37,0.33,0.48,0.45,0.46,0.49,,0.97, 770
GPT3 Babbage,$1 \times 10^5$,0.45,0.36,0.33,0.43,0.45,0.45,0.47,,\textbf{0.98}, 6700
eXpose + T5 Labels,$1 \times 10^7$,0.45,0.32,0.27,0.45,0.43,0.45,0.49,0.42,0.98, 3.3
BERTiny + T5 Labels,$1 \times 10^7$,0.46,0.35,0.30,0.47,0.44,0.46,0.5,0.57,0.98, 4.4
URLTran + T5 Labels,$1 \times 10^7$,\textbf{0.47},0.38,0.34,0.50,0.46,0.47,0.5,14.29,\textbf{0.99}, 110
\end{filecontents*}
\csvreader[
    tabular=l r r r r r,
    table head=\hline \textbf{Model} & 
    \begin{tabular}{@{}c@{}}\rule{0pt}{2.5ex}\textbf{Accuracy} \\ \textbf{Time and Domain Split} \\ \rule{0pt}{1.5ex}\end{tabular} & 
    \begin{tabular}{@{}c@{}}\rule{0pt}{2.5ex}\textbf{Accuracy} \\ \textbf{Time Split} \\ \rule{0pt}{1.5ex}\end{tabular} &
    \begin{tabular}{@{}c@{}}\rule{0pt}{2.5ex}\textbf{Parameter Count} \\ \textbf{in millions} \\ \rule{0pt}{1.5ex}\end{tabular} & 
    \begin{tabular}{@{}c@{}}\rule{0pt}{2.5ex}\textbf{Parameter Count Relative} \\ \textbf{to the Teacher (\%)} \\ \rule{0pt}{1.5ex}\end{tabular} & \textbf{Training Samples Count} \\
\midrule
]{main_table.csv}{}{
    \csvcoli & 
    \csvcoliii & 
    \csvcolxi &
    \csvcolxii & 
    \csvcolx & 
    \csvcolii 
    \ifnumequal{\thecsvinputline}{6}{ \\ \midrule}{} 
}
}
\end{table}

\subsection{Discussion}

A comparison of the models' performance on the two evaluation splits reveals that results on the time split, the traditional validation approach, are overly optimistic. Small models such as BERTiny, trained merely on signature-driven data, exhibit performance comparable to T5 and GPT-3. The disparity in model performance between the domain and time split versus the time split, particularly for small models, underscores that signature-sourced data is repetitive and can be memorized with just a few million parameters. Time split validation measures a model's ability to match the signature distribution while in a production setting the primary concern within the context of the overall pipeline is a model's capacity to generalize to new data from the long tail that falls outside the coverage of signatures.

When considering the domain and time split—which aligns more closely with real-world performance on unlabeled data—small models no longer match the performance of LLMs, as seen in Figure \ref{fig:results_scaling}. Beyond 10,000 samples, LLMs show minimal to no performance gains when scaling up further. Conversely, the performance of small models and the baseline has not yet converged at 5 million training samples. This demonstrates the sample-efficiency of LLMs in the domain of website content categorization. 

\begin{figure}[!h]
    \centering
    \begin{minipage}[t]{0.45\textwidth}
        \vspace{0pt}
        \includegraphics[width=\textwidth]{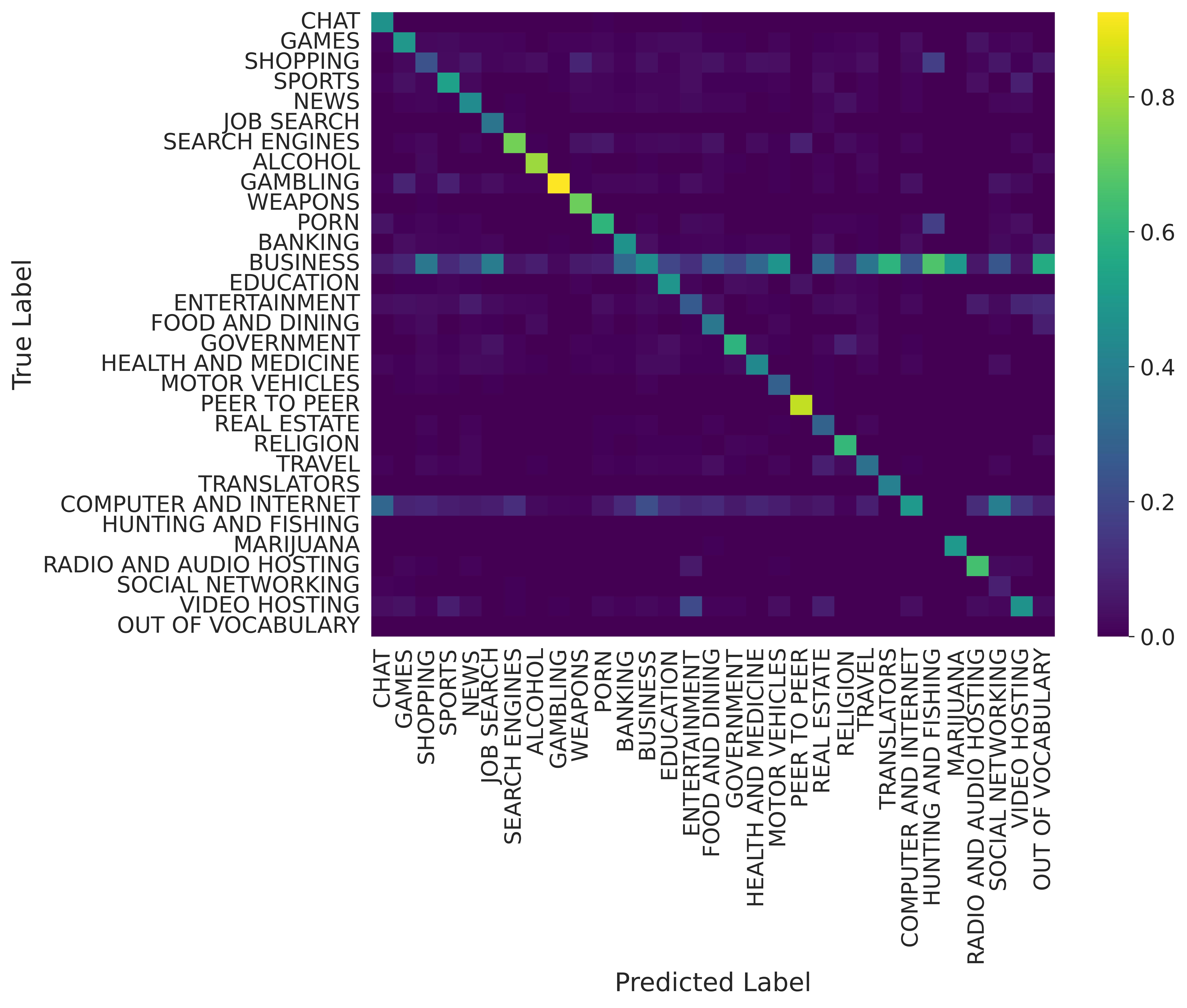}
        \caption{Normalized confusion matrix on the domain and time split for the best T5 Large configuration}
        \label{fig:confusion_matrix}
    \end{minipage}
    \hfill
    \begin{minipage}[t]{0.45\textwidth}
        \vspace{0pt}
        \centering
        \normalsize 
        \renewcommand{\arraystretch}{2.3} 
        \resizebox{\textwidth}{!}{%
            \begin{tabular}{lll}
            \toprule
            \textbf{Domain} & \textbf{LLM Label} & \textbf{True Label} \\
            \midrule
            citytocoastneurosurgery.com.au & HEALTH AND MEDICINE & BUSINESS  \\
            twittodon.com & SOCIAL NETWORKING & COMPUTER AND INTERNET \\
            robinsonmalls.com/mall-info & SHOPPING & BUSINESS \\
            
            online-weinshop.at & SHOPPING & ALCOHOL  \\
            www.fourbakery.com & BUSINESS & FOOD \\

            sargenttoolsonline.com & BUSINESS & SHOPPING  \\

            \noalign{\vskip 1pt}
            \cline{1-3}
            \noalign{\vskip 1pt}

            praeyforthegods.com & RELIGION & GAMES  \\
            www.hygiene-3d.com & HEALTH AND MEDICINE & SHOPPING \\

            \noalign{\vskip 1pt}
            \cline{1-3}
            \noalign{\vskip 1pt}
            
            beta.x9zb.live  & COMPUTING AND INTERNET & GAMBLING  \\
            www.857zb6.com  & ENTERTAINMENT & SPORTS \\
            www.lxf.cz  & BUSINESS & SHOPPING  \\
            g11.178tiyu.com  & ENTERTAINMENT & SPORTS \\
            
            \bottomrule
            \end{tabular}%
        }
        \captionof{table}{Examples of LLM (T5) misclassifications. Comparison of LLM performance on the domain and time split, highlighting the impact of the single-label strategy and keyword-absent URLs.}
        \label{tab:FPs}
    \end{minipage}
\end{figure}

LLMs outperform student models in terms of performance, but they still fall short of perfection when applied to domain and test splits. This discrepancy can be attributed to two main factors.
First, due to dataset limitations, web content classification is framed as a single-label classification problem. Table \ref{tab:FPs} displays a set of LLM misclassifications on domain and time split, highlighting that a URL could potentially belong to multiple categories. In the first three samples, the analyst opted for the more generic label, while the model choose the more generic labels in the following three samples. Both predicted and true labels could be considered correct in all six cases, suggesting that the true performance is likely better than the metrics indicate because of the single-label limitation. This trade-off between equally correct specific and general labels becomes evident when examining the confusion matrix in Figure \ref{fig:confusion_matrix} for a T5 Large model's performance on the domain and time split. As we can see on the confusion matrix, the LLM tends to generate class labels that are more specific than manual labels.

The second factor occurs when a URL lacks keywords or context related to its category, as demonstrated by the last six entries in Table \ref{tab:FPs}. For the middle two URLs, the model was misled by a prominent keyword in the URL, which was unrelated to its content. The final four URLs contain no apparent signal. Consequently, the large-scale pre-training of LLMs struggles to effectively transfer knowledge to a URL from the long tail. This means that if a URL lacks clear or strong indicators of its category, the LLM may not accurately classify it, leading to misclassifications.

Our results reveal that mixing in the LLM-generated labels significantly enhances the performance of student models BERTiny and eXpose as we can see on Figure \ref{fig:results_augmentation}. Through this simple form of augmentation, we nearly matched the 46.3\% accuracy of T5 Large, the best-performing LLM, with a transformer model that has a parameter count several orders of magnitude smaller (0.57\% of the teacher) and could reasonably be deployed in-line in production.

\section{Conclusion}

In conclusion, our paper contributes to the field of web content classification with the development of lightweight models distilled from fine-tuned LLMs. We have demonstrated that LLMs, when fine-tuned on data labeled with domain propagation signatures, significantly outperform the current state-of-the-art approach on the long tail categorization problem. Our teacher-student training approach enables the distillation of LLMs into models 175 times smaller without sacrificing accuracy, thus making deployment practical in a wide variety of new contexts. The amount of manual labels required to finetune the teacher LLM is orders of magnitude smaller than what is required for convergence of the current state-of-the-art approach. Furthermore, we have proposed a new validation approach that better measures model performance in more realistic scenarios, which should be adapted by the community to improve generalization capabilities to unseen data.

Expanding beyond web content classification, the cybersecurity field could greatly benefit from proven methods of distilling large language models (LLMs) into more compact versions. This approach is particularly valuable when dealing with large data volumes and expensive training samples, especially when the model is applied to out-of-distribution cases. For addressing web content classification tasks specifically, we suggest future work should focus on augmenting the training data and feature space with HTML and image data, utilizing GPT-4 as a teacher, allowing URLs to have more than one label, and re-working signatures for the assignment of general categories. 

\bibliographystyle{unsrt}  
\bibliography{references}  

\newpage

\appendix 
\label{app:A}

\setcounter{table}{0}
\setcounter{figure}{0}

\section{Appendix}
\renewcommand{\thetable}{\Alph{section}\arabic{table}}
\renewcommand{\thefigure}{\Alph{section}\arabic{figure}}

\FloatBarrier

\begin{figure}[ht]
    \centering
    \includegraphics[width=0.85\textwidth]
    {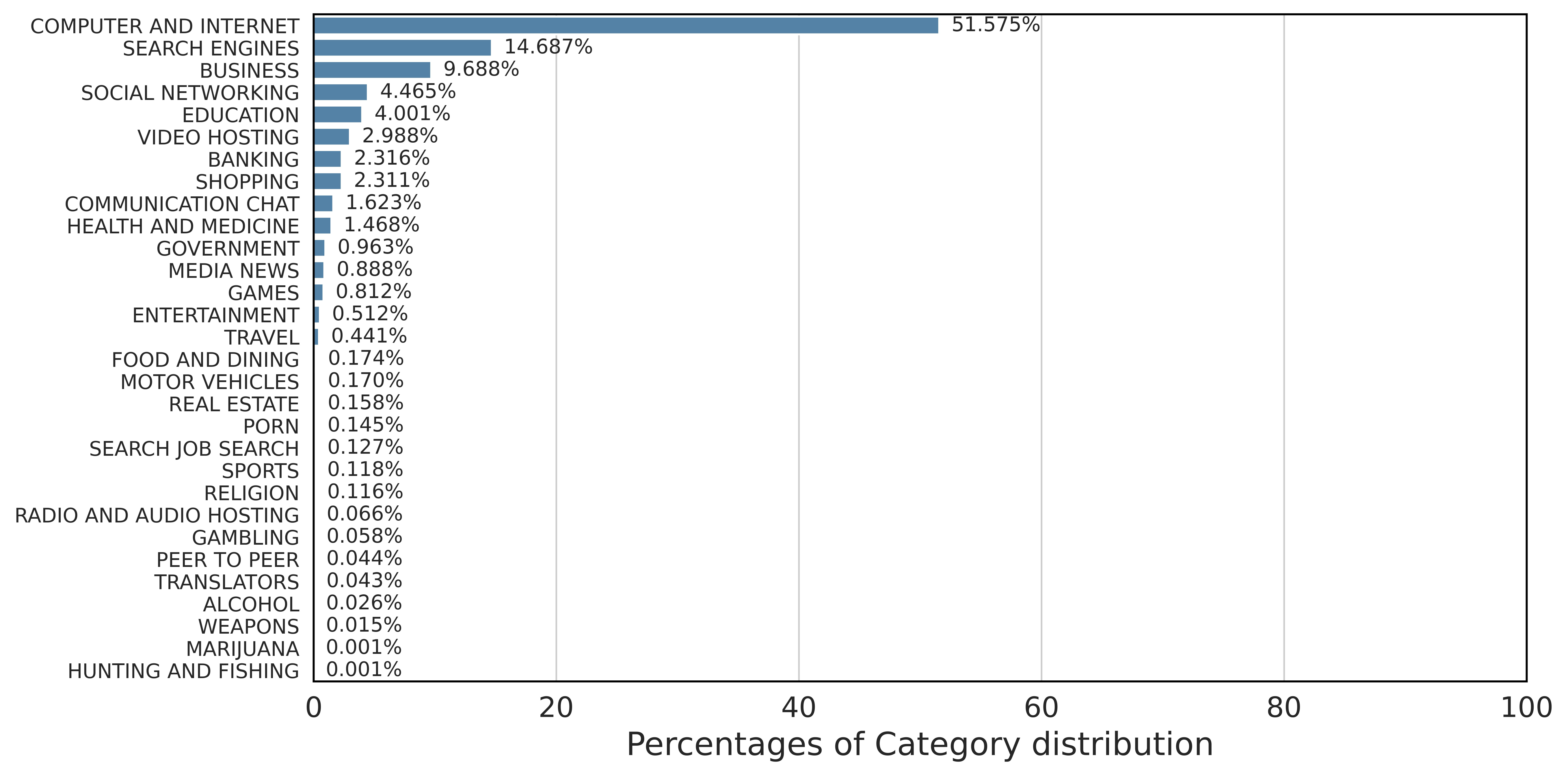}
    \caption{Class Distribution. Displaying the 30 classes and their distribution over the datasets.}
    \label{fig:category_distribution}
\end{figure}

\vspace{3 cm}

\begin{table}
\centering
\scriptsize
\caption{Model performance on Domain and Time Split}
\label{tab:model-performance-domain}
\makebox[\linewidth]{
\begin{filecontents*}{appendix_table_domain.csv}
model,n_samples_int,accuracy,macro_f1,macro_recall,macro_precision,weighted_f1,weighted_recall,weighted_precision
BERTiny,10,0.051,0.045,0.077,0.06,0.055,0.051,0.207
BERTiny,100,0.089,0.083,0.152,0.089,0.081,0.089,0.242
BERTiny,1000,0.117,0.079,0.123,0.119,0.11,0.117,0.285
BERTiny,10000,0.251,0.098,0.109,0.137,0.24,0.251,0.297
BERTiny,100000,0.348,0.208,0.185,0.317,0.344,0.348,0.39
BERTiny,1000000,0.333,0.213,0.196,0.279,0.332,0.333,0.366
BERTiny,2500000,0.348,0.232,0.208,0.313,0.347,0.348,0.38
BERTiny,5000000,0.357,0.241,0.221,0.304,0.357,0.357,0.384
BERTiny + LLM Labels (0.0 Ratio),10000000,0.375,0.253,0.23,0.335,0.372,0.375,0.405
BERTiny + LLM Labels (0.25 Ratio),10000000,0.457,0.34,0.296,0.456,0.437,0.457,0.494
BERTiny + LLM Labels (0.5 Ratio),10000000,0.456,0.336,0.293,0.452,0.436,0.456,0.492
BERTiny + LLM Labels (0.75 Ratio),10000000,0.449,0.323,0.279,0.442,0.427,0.449,0.482
BERTiny + LLM Labels (1.0 Ratio),10000000,0.462,0.347,0.299,0.473,0.438,0.462,0.5
GPT-3 Babbage,10,0.267,0.235,0.362,0.214,0.265,0.267,0.418
GPT-3 Babbage,100,0.316,0.265,0.384,0.232,0.317,0.316,0.41
GPT-3 Babbage,1000,0.375,0.315,0.379,0.294,0.382,0.375,0.434
GPT-3 Babbage,10000,0.444,0.364,0.354,0.402,0.45,0.444,0.471
GPT-3 Babbage,100000,0.449,0.36,0.328,0.428,0.45,0.449,0.471
GPT-3 Babbage,1000000,0.444,0.364,0.33,0.465,0.445,0.444,0.469
T5 Large,10,0.211,0.254,0.284,0.324,0.244,0.211,0.415
T5 Large,100,0.273,0.277,0.374,0.282,0.278,0.273,0.402
T5 Large,1000,0.379,0.34,0.382,0.367,0.362,0.379,0.466
T5 Large,10000,0.463,0.368,0.332,0.477,0.452,0.463,0.489
T5 Large,100000,0.427,0.372,0.353,0.426,0.43,0.427,0.476
T5 Large,1000000,0.428,0.296,0.256,0.434,0.404,0.428,0.455
T5 Large,5000000,0.438,0.329,0.286,0.454,0.435,0.438,0.471
URLTran (BERT),10,0.135,0.055,0.092,0.096,0.141,0.135,0.244
URLTran (BERT),100,0.186,0.176,0.285,0.168,0.18,0.186,0.336
URLTran (BERT),1000,0.304,0.261,0.308,0.261,0.314,0.304,0.401
URLTran (BERT),10000,0.379,0.278,0.266,0.322,0.384,0.379,0.412
URLTran (BERT),100000,0.383,0.246,0.224,0.325,0.377,0.383,0.41
URLTran (BERT),1000000,0.358,0.264,0.254,0.333,0.362,0.358,0.403
URLTran (BERT),5000000,0.377,0.288,0.278,0.357,0.382,0.377,0.435
URLTran + LLM Labels (0.0 Ratio),10000000,0.416,0.317,0.307,0.364,0.421,0.416,0.446
URLTran + LLM Labels (0.25 Ratio),10000000,0.465,0.357,0.326,0.446,0.454,0.465,0.493
URLTran + LLM Labels (0.5 Ratio),10000000,0.465,0.364,0.338,0.443,0.454,0.465,0.498
URLTran + LLM Labels (0.75 Ratio),10000000,0.466,0.364,0.329,0.453,0.452,0.466,0.499
URLTran + LLM Labels (1.0 Ratio),10000000,0.468,0.375,0.338,0.496,0.455,0.468,0.497
eXpose (Conv),10,0.033,0.023,0.05,0.049,0.043,0.033,0.257
eXpose (Conv),100,0.062,0.052,0.107,0.062,0.061,0.062,0.208
eXpose (Conv),1000,0.097,0.059,0.093,0.074,0.094,0.097,0.205
eXpose (Conv),10000,0.223,0.051,0.063,0.105,0.177,0.223,0.255
eXpose (Conv),100000,0.256,0.042,0.049,0.129,0.187,0.256,0.264
eXpose (Conv),1000000,0.262,0.107,0.103,0.175,0.238,0.262,0.279
eXpose (Conv),5000000,0.302,0.153,0.143,0.229,0.291,0.302,0.335
eXpose + LLM Labels (0.0 Ratio),10000000,0.315,0.168,0.157,0.237,0.306,0.315,0.352
eXpose + LLM Labels (0.25 Ratio),10000000,0.444,0.316,0.266,0.45,0.418,0.444,0.481
eXpose + LLM Labels (0.5 Ratio),10000000,0.436,0.308,0.259,0.445,0.415,0.436,0.471
eXpose + LLM Labels (0.75 Ratio),10000000,0.422,0.28,0.231,0.426,0.4,0.422,0.458
eXpose + LLM Labels (1.0 Ratio),10000000,0.45,0.318,0.268,0.453,0.425,0.45,0.488
\end{filecontents*}
\csvreader[
    tabular=ccccccccc,
    table head=\textbf{Model} & \textbf{Samples} & \textbf{Accuracy} & \textbf{Macro F1} & \textbf{Macro Recall} & \textbf{Macro Precision} & \textbf{Weighted F1} & \textbf{Weighted Recall} & \textbf{Weighted Precision} \\\hline,
    late after line=\\
]{appendix_table_domain.csv}{}{\csvlinetotablerow}
}
\end{table}

\begin{table}
\centering
\scriptsize
\caption{Model performance on Time Split}
\label{tab:model-performance-time}
\makebox[\linewidth]{
\begin{filecontents*}{appendix_table_time.csv}
model,n_samples_int,accuracy,macro_f1,macro_recall,macro_precision,weighted_f1,weighted_recall,weighted_precision
BERTiny,10,0.206,0.104,0.27,0.128,0.226,0.206,0.688
BERTiny,100,0.307,0.155,0.401,0.15,0.229,0.307,0.551
BERTiny,1000,0.355,0.135,0.301,0.176,0.314,0.355,0.646
BERTiny,10000,0.432,0.177,0.228,0.197,0.44,0.432,0.58
BERTiny,100000,0.912,0.713,0.722,0.714,0.915,0.912,0.92
BERTiny,1000000,0.967,0.78,0.765,0.802,0.966,0.967,0.966
BERTiny,2500000,0.973,0.821,0.804,0.844,0.973,0.973,0.973
BERTiny,5000000,0.976,0.856,0.838,0.882,0.976,0.976,0.976
BERTiny + LLM Labels (0.0 Ratio),10000000,0.987,0.867,0.858,0.878,0.987,0.987,0.987
BERTiny + LLM Labels (0.25 Ratio),10000000,0.971,0.835,0.814,0.86,0.972,0.971,0.974
BERTiny + LLM Labels (0.5 Ratio),10000000,0.98,0.858,0.841,0.878,0.98,0.98,0.98
BERTiny + LLM Labels (0.75 Ratio),10000000,0.982,0.877,0.859,0.902,0.982,0.982,0.983
BERTiny + LLM Labels (1.0 Ratio),10000000,0.608,0.546,0.546,0.605,0.647,0.608,0.763
GPT-3 Babbage,10,0.429,0.304,0.593,0.269,0.369,0.429,0.587
GPT-3 Babbage,100,0.603,0.409,0.716,0.343,0.597,0.603,0.745
GPT-3 Babbage,1000,0.771,0.62,0.793,0.534,0.771,0.771,0.81
GPT-3 Babbage,10000,0.969,0.818,0.831,0.809,0.97,0.969,0.971
GPT-3 Babbage,100000,0.973,0.837,0.829,0.849,0.974,0.973,0.974
GPT-3 Babbage,1000000,0.984,0.855,0.839,0.874,0.984,0.984,0.984
T5 Large,10,0.394,0.35,0.497,0.344,0.391,0.394,0.452
T5 Large,100,0.475,0.375,0.61,0.328,0.432,0.475,0.607
T5 Large,1000,0.688,0.542,0.715,0.489,0.673,0.688,0.738
T5 Large,10000,0.927,0.794,0.796,0.8,0.932,0.927,0.942
T5 Large,100000,0.959,0.821,0.824,0.821,0.959,0.959,0.96
T5 Large,1000000,0.966,0.772,0.73,0.838,0.966,0.966,0.968
T5 Large,5000000,0.975,0.835,0.807,0.874,0.975,0.975,0.975
URLTran (BERT),10,0.122,0.093,0.192,0.166,0.134,0.122,0.459
URLTran (BERT),100,0.396,0.272,0.598,0.238,0.325,0.396,0.641
URLTran (BERT),1000,0.636,0.508,0.773,0.432,0.63,0.636,0.753
URLTran (BERT),10000,0.944,0.757,0.812,0.72,0.945,0.944,0.949
URLTran (BERT),100000,0.947,0.778,0.791,0.774,0.949,0.947,0.954
URLTran (BERT),1000000,0.971,0.817,0.806,0.836,0.971,0.971,0.972
URLTran (BERT),5000000,0.956,0.82,0.802,0.873,0.953,0.956,0.957
URLTran + LLM Labels (0.0 Ratio),10000000,0.992,0.914,0.897,0.94,0.992,0.992,0.992
URLTran + LLM Labels (0.25 Ratio),10000000,0.985,0.859,0.851,0.869,0.985,0.985,0.985
URLTran + LLM Labels (0.5 Ratio),10000000,0.988,0.889,0.884,0.896,0.988,0.988,0.988
URLTran + LLM Labels (0.75 Ratio),10000000,0.991,0.908,0.896,0.927,0.991,0.991,0.991
URLTran + LLM Labels (1.0 Ratio),10000000,0.74,0.676,0.672,0.706,0.747,0.74,0.826
eXpose (Conv),10,0.125,0.075,0.165,0.116,0.174,0.125,0.707
eXpose (Conv),100,0.296,0.168,0.362,0.164,0.279,0.296,0.636
eXpose (Conv),1000,0.306,0.193,0.327,0.215,0.359,0.306,0.693
eXpose (Conv),10000,0.565,0.18,0.204,0.26,0.543,0.565,0.652
eXpose (Conv),100000,0.549,0.156,0.147,0.399,0.523,0.549,0.611
eXpose (Conv),1000000,0.904,0.539,0.478,0.744,0.9,0.904,0.906
eXpose (Conv),5000000,0.928,0.662,0.601,0.784,0.925,0.928,0.927
eXpose + LLM Labels (0.0 Ratio),10000000,0.978,0.838,0.809,0.878,0.977,0.978,0.977
eXpose + LLM Labels (0.25 Ratio),10000000,0.964,0.78,0.748,0.823,0.965,0.964,0.968
eXpose + LLM Labels (0.5 Ratio),10000000,0.971,0.823,0.796,0.858,0.972,0.971,0.973
eXpose + LLM Labels (0.75 Ratio),10000000,0.976,0.842,0.811,0.886,0.976,0.976,0.977
eXpose + LLM Labels (1.0 Ratio),10000000,0.49,0.413,0.397,0.477,0.552,0.49,0.745
\end{filecontents*}
\csvreader[
    tabular=ccccccccc,
    table head=\textbf{Model} & \textbf{Samples} & \textbf{Accuracy} & \textbf{Macro F1} & \textbf{Macro Recall} & \textbf{Macro Precision} & \textbf{Weighted F1} & \textbf{Weighted Recall} & \textbf{Weighted Precision} \\\hline,
    late after line=\\
]{appendix_table_time.csv}{}{\csvlinetotablerow}
}
\end{table}

\begin{table}
  \centering
  \caption{BERTiny Parameters}
    \begin{tabular}{lll}
    \toprule
    \textbf{Parameter} & \textbf{Value} \\
    \midrule
    model\textunderscore name &  prajjwal1/bert-tiny \\
hidden\textunderscore size &  128 \\
hidden\textunderscore act &  gelu \\
initializer\textunderscore range &  0.02 \\ 
vocab\textunderscore size &  30522 \\
hidden\textunderscore dropout\textunderscore prob &  0.1 \\
num\textunderscore attention\textunderscore heads &  2 \\
type\textunderscore vocab\textunderscore size &  2 \\
max\textunderscore position\textunderscore embeddings &  512 \\
num\textunderscore hidden\textunderscore layers &  2 \\
intermediate\textunderscore size &  512 \\
attention\textunderscore probs\textunderscore dropout\textunderscore prob &  0.1 \\
maximum sequence length &  128 \\
learning rate &  1e-4 \\
batch size &  49160 \\
optimizer &  Adam \\
maximum training epochs &  20 \\
    \bottomrule
    \end{tabular}%
  \label{tab:bertiny_hyperparameters}%
\end{table}%

\begin{table}
  \centering
  \caption{BERT Parameters}
    \begin{tabular}{lll}
    \toprule
    \textbf{Parameter} & \textbf{Value} \\
    \midrule
model\textunderscore name &  bert-base-uncased \\
hidden\textunderscore size &  768 \\
hidden\textunderscore act &  gelu \\
initializer\textunderscore range &  0.02 \\
vocab\textunderscore size &  30522 \\
hidden\textunderscore dropout\textunderscore prob &  0.1 \\
num\textunderscore attention\textunderscore heads &  12 \\
type\textunderscore vocab\textunderscore size &  2 \\
max\textunderscore position\textunderscore embeddings &  512 \\
num\textunderscore hidden\textunderscore layers &  12 \\
intermediate\textunderscore size &  3072 \\
attention\textunderscore probs\textunderscore dropout\textunderscore prob &  0.1 \\
maximum sequence length &  128 \\
learning rate &  1e-4 \\
batch size &  2048 \\
optimizer &  Adam \\
maximum training epochs &  20 \\
    \bottomrule
    \end{tabular}%
  \label{tab:bert_hyperparameters}%
\end{table}%

\begin{table}
  \centering
  \caption{eXpose Parameters}
    \begin{tabular}{lll}
    \toprule
    \textbf{Parameter} & \textbf{Value} \\
    \midrule
vocab\textunderscore size &  76 \\
filter\textunderscore size &  128 \\
dropout &  0.05 \\
learning rate &  1e-3 \\
batch size &  49160 \\
optimizer &  Adam \\
maximum training epochs &  20 \\
    \bottomrule
    \end{tabular}%
  \label{tab:eXpose_hyperparameters}%
\end{table}%

\begin{table}
  \centering
  \caption{T5 Parameters}
    \begin{tabular}{lll}
    \toprule
    \textbf{Parameter} & \textbf{Value} \\
    \midrule
model\textunderscore name &  t5-large \\
d\textunderscore ff &  4096 \\
d\textunderscore kv &  64 \\
d\textunderscore model &  1024 \\
dropout\textunderscore rate &  0.1 \\
initializer\textunderscore facto &  1.0 \\
layer\textunderscore norm\textunderscore epsilon &  1e-06 \\
n\textunderscore positions &  512 \\
num\textunderscore heads &  16 \\
num\textunderscore layers &  24 \\
relative\textunderscore attention\textunderscore num\textunderscore buckets &  32 \\
vocab\textunderscore size &  32128 \\
maximum sequence length & 128 \\
learning rate &  1e-3 \\
batch size &  60 \\
optimizer &  Adafactor \\
Adafactor scale\textunderscore parameter &  False \\
Adafactor relative\textunderscore step & False  \\
Adafactor warmup\textunderscore init &  False \\
maximum training epochs &  15 \\
    \bottomrule
    \end{tabular}%
  \label{tab:t5_hyperparameters}%
\end{table}%

 \begin{table}
   \centering
   \caption{Adafactor Parameters}
     \begin{tabular}{lll}
     \toprule
     \textbf{Parameter} & \textbf{Value} \\
     \midrule
     scale\textunderscore parameter &  False \\
     relative\textunderscore step & False  \\
     warmup\textunderscore init &  False \\
     \bottomrule
     \end{tabular}%
   \label{tab:adafactor_hyperparameters}%
 \end{table}%

\FloatBarrier 


\end{document}